# Binarizer at SemEval-2018 Task 3: Parsing dependency and deep learning for irony detection

**Nishant Nikhil**
IIT Kharagpur
Kharagpur, India
nishantnikhil@iitkgp.ac.in

**Muktabh Mayank Srivastava**
ParallelDots, Inc.
muktabh@paralleldots.com

## Abstract

In this paper, we describe the system submitted for the SemEval 2018 Task 3 (Irony detection in English tweets) Subtask A by the team Binarizer. Irony detection is a key task for many natural language processing works. Our method treats ironical tweets to consist of smaller parts containing different emotions. We break down tweets into separate phrases using a dependency parser. We then embed those phrases using an LSTM-based neural network model which is pre-trained to predict emoticons for tweets. Finally, we train a fully-connected network to achieve classification.

## 1 Introduction

The micro-blogging site Twitter has created an abundance of data about opinions and sentiments regarding almost every aspect of daily life. A deeper study of the public opinion can be obtained by applying natural language processing techniques on this data. However, the performance of these NLP models is detrimentally affected by irony (Pozzi et al., 2016). As per the Oxford English Dictionary, irony is the expression of one's meaning by using language that normally signifies the opposite, typically for humorous or emphatic effect. This deviation between what is said and what is intended makes irony hard to detect. Being a platform where users are free to communicate and express themselves colloquially, Twitter generates considerable data injected with irony. Studying this would provide us with a better sentiment analysis of these tweets.

Prior work on irony detection includes the use of unigrams and emoticons (González-Ibánez et al., 2011; Carvalho et al., 2009; Barbieri et al., 2014). Maynard and Greenwood (2014) describe an unsupervised pattern mining approach where the sentiment of the hashtag in the tweet is proposed to be a key indicator of sarcasm. If the sentiment of the tweet does not match the sentiment of the hashtag, it is predicted to be sarcastic. Riloff et al. (2013) illustrates a semi-supervised approach where rule-based classifiers are used to look for negative situation phrases and positive verbs in a sentence. Tsur et al. (2010) build pattern-based features that detect the presence of discriminative patterns as extracted from a large sarcasm-labelled corpus. N-gram-based approaches have also been used (Davidov et al., 2010; Ptáček et al., 2014; Joshi et al., 2015) with sentiment features. Joshi et al. (2017) use similarity between word embeddings as feature and Poria et al. (2016) use convolutional neural networks to extract sentiment, emotion and personality features for sarcasm detection.

SemEval-2018 Task 3 (the 12th workshop on semantic evaluation) specifies two subtasks in relation to irony detection in English tweets (Van Hee et al., 2018). In subtask A the goal was to train a binary classifier that detects whether a given tweet is ironic or not. Subtask B was a multi-class classification problem where four labels were specified to describe the nature of irony (verbal irony by means of a polarity contrast, situational irony, other verbal irony, and non-ironic). The goal was to assign one of the four labels to each tweet.

We propose a new method which considers ironical tweets to be collections of smaller parts containing different emotions. We break down tweets into these collections using a dependency parser and embed them using DeepMoji (Felbo et al., 2017) which is pre-trained to predict emoticons for tweets. Finally we train a classifier to detect irony. The paper is organized as follows: We discuss our methods in section 2. Section 3 contains the details about the experiments and the training data. In Section 4 we discuss the results and Section 5 concludes the paper with closing remarks.

## 2 Method

In order to identify the chunks of various emotions in an ironic tweet, we split the tweets into phrases using a dependency parser. We use Tweeboparser (Kong et al., 2014), which is a dependency parser for English tweets. The parser is trained on a subset of a labelled corpus for 929 tweets (12,318 tokens) drawn from the POS-tagged tweet corpus of Owoputi et al. (2013), Tweebank. TweeboParser predicts the syntactic structure of the tweet represented by unlabelled dependencies. Tweets contain multiple sentences or fragments called "utterances" each with their own syntactic root disconnected from the others. Since a tweet often contains more than one utterance, the output of TweeboParser will often be a multi-rooted graph over the tweet. Also, many elements in tweets have no syntactic function. These include, in many cases, hashtags, URLs, and emoticons. TweeboParser attempts to exclude these tokens from the parse tree. For our purpose, we club the words arising from the same root to create a phrase. Multiple roots would create multiple phrases. As we can see from Figure 1, these phrases can convey the different sentiments attached to the different subjects of the tweet.

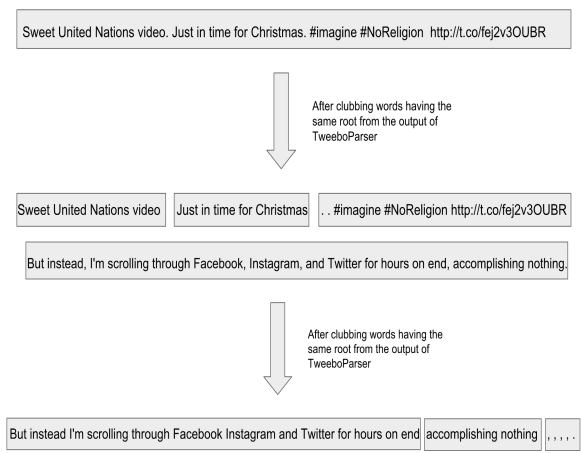

Figure 1: Parser results

After extracting a set of phrases for the sentence, we embed the phrases into vectors. We used the DeepMoji (Felbo et al., 2017) model, which is trained on 1.2 billion tweets with emojis to understand how language is used to express emotions. It encodes the provided phrase into a 2,304-dimensional feature vector. Under the hood, DeepMoji model projects each word into a 256-dimensional vector space followed by a hyperbolic tangent activation function. After this, two bidirectional LSTMs with 1,024 hidden units each are used to capture the context of each word. Finally, the model uses skip connections from each layer to an attention layer and hence the attention layer outputs a 2,304 (256+1,024+1,024) dimensional vector. Now this 2,304-dimensional output is connected to a softmax layer for classification. We did not use the final softmax layer but took the 2,304-dimensional vector for each phrase. As the model was trained for prediction of emoticons, this feature vector contains information about the semantic and sentimental content of the phrases. To make the predictions we need to account for the sentiment behind every utterance. To this end, we concatenate these vectors and pass the resulting concatenated vector through a fully-connected network as described in Figure 2.

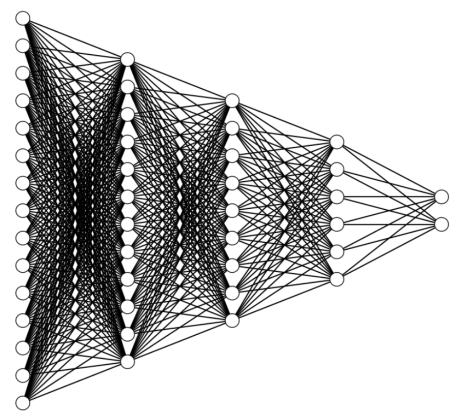

Figure 2: Neural network architecture

Tweets can have a varying number of roots, which implies that they split into a varying number of phrases. Our model considers a maximum of nine roots. A tweet with an excess of nine roots is truncated suitably. On the other hand, a tweet with less than nine roots is zero-padded. We have described the complete process flowchart in Figure 3.

## 3 Experiments

For subtask A, we were provided with a dataset consisting of tweets along with a binary class (0 or 1) which indicates whether this tweet is ironic or not (0 for non-ironic tweets and 1 for ironic tweets). The data was collected from Twitter API by querying tweets using the hashtags #irony, #sarcasm and #not, with subsequent manual an-notation to remove noise. 3,833 tweets for training and 784 tweets for testing were provided. The evaluation was done by using accuracy, precision, recall and F1 score.

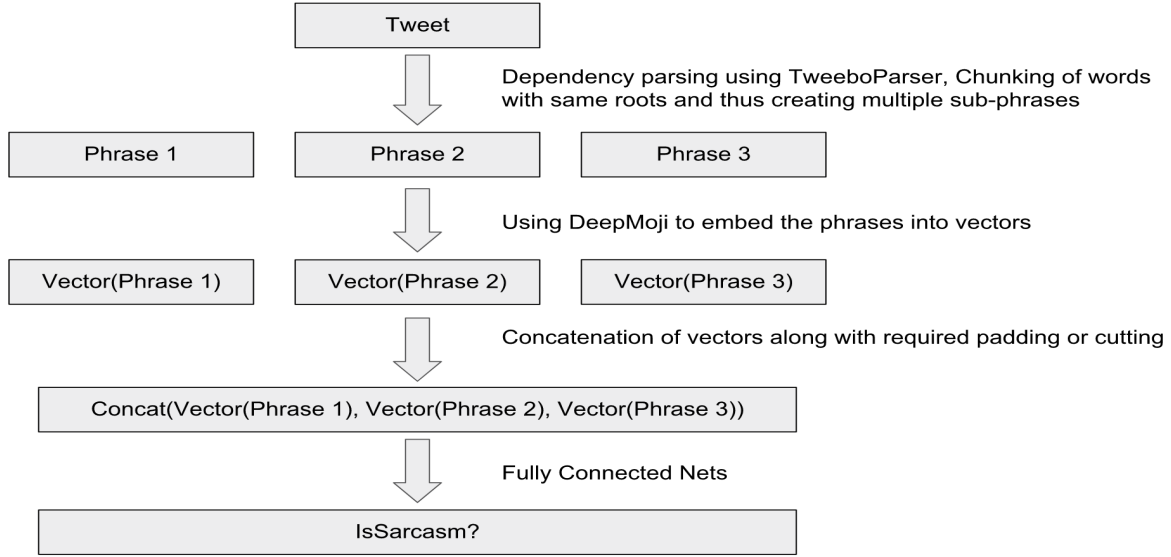

Figure 3: Process Flowchart

$$Accuracy : \frac{\text{number of correctly predicted instances}}{\text{total number of instances}}$$

$$Precision : \frac{\text{number of correctly predicted instances}}{\text{number of predicted labels}}$$

$$Recall : \frac{\text{number of correctly predicted instances}}{\text{number of labels in the gold standard}}$$

$$F1\ score : \frac{2 \text{ x precision x recall}}{\text{precision + recall}}$$

We used the pipeline described in Figure 3. The final step of the process used a fully connected neural network with four layers. The input layer of the FC network has a dimension of 20,736 (2,304*9), the second layer has a dimension of 9,216 (2,304*4), the third layer has a dimension of 2,304 and the fourth layer has 256 dimensions. The final layer has 2 dimensions, with one for each class. This is depicted in Figure 2. We used the hyperbolic tangent activation function in all of the layers, and stochastic gradient descent with a learning rate of 0.01 and a momentum of 0.5.

Two models were then devised. The difference in these models lies in the input supplied to the FC network. In the first model, this input is the concatenation of the vectors obtained by embedding phrases. In the second model, the input is the concatenation of the input in the first model along with a 2,304-dimensional vector representing the embedding of the tweet as a whole. The results we get from various experiments on these models are shown in Table 1.

| Method | Accuracy | Precision | Recall | F1 score |
|---|---|---|---|---|
| Winning team | 0.7347 | 0.6304 | 0.8006 | 0.7054 |
| Our System $\beta 1$ | **0.6659** | **0.5527** | **0.6471** | **0.5962** |
| Our System $\beta 2$ | 0.6390 | 0.5198 | 0.6941 | 0.5944 |
| Our System $\alpha$ | 0.6951 | 0.6197 | 0.5176 | 0.5641 |

Table 1: Results SemEval Task 3A

System $\alpha$ is the first model. The best F1 score for this model was achieved after four epochs, as shown in Table 1. System $\beta 1$ and $\beta 2$ are the second model running for five and four epochs respectively.

## 4 Results and Discussions

We participated only in the shared task 3A as the team Binarizer. We came ninth as per accuracy and seventeenth as per F1 score among the forty-three participating systems. Due to a glitch on our side during submission the results are based on 446 out of 784 instances in the test data.

The models perform better than the baseline system as per the competition leaderboard. This reinforces the notion that separate phrases in a tweet carry information required for irony detection. System $\alpha$ has greater precision whereas System $\beta$ has higher recall. So an application which demands urgent detection of ironic tweets would profit more from System $\beta$. This demonstrates that the sentiment information of the context provided from the whole tweet is also important.

## 5 Conclusion and Future works

We have shown how using the sentiments of different segments of tweets can enable irony detection. From the results of our experiments, we conclude that the segments have sufficient sentiment information in them for the identification of irony. In future research, we aim to improve the algorithm for parsing these chunks by replacing the dependency parser. Also, more experimentation can be performed for the last part of the pipeline. As the phrases from the tweets are sequences themselves, we can apply sequence modelling with LSTMs or CNNs.